\documentclass{article}

\pdfoutput=1 
\usepackage[accepted]{icml2007}
\usepackage{amsmath}
\usepackage{theorem}
\usepackage{amssymb}
\usepackage{amsmath}
\usepackage{graphicx}
\usepackage{algorithm}
\usepackage{algorithmic}
\usepackage{url}
\usepackage{subfigure}
\usepackage{mdwlist}
\usepackage{multirow}
\usepackage[small,compact]{titlesec}
\usepackage[small]{caption}
\usepackage{paralist}
\usepackage{graphicx}
\usepackage{natbib}

\renewenvironment{thebibliography}[1]{%
  \begin{oldthebibliography}{#1}%
    \setlength{\parskip}{0ex}%
    \setlength{\itemsep}{0ex}%
  }%
  {%
  \end{oldthebibliography}%
}

\newcommand{\twoco}[1]{\multicolumn{2}{c|}{#1}}

\icmltitlerunning{Supervised Feature Selection via Dependence
Estimation}

\input{Definitions}

\begin{document}

\twocolumn[\icmltitle{Supervised Feature Selection via Dependence
Estimation}
\icmlauthor{Le Song}{lesong@it.usyd.edu.au}
\icmladdress{NICTA, Statistical Machine Learning Program,
Canberra, ACT 0200, Australia; and University of Sydney}
\icmlauthor{Alex Smola}{alex.smola@gmail.com} \icmladdress{NICTA,
Statistical Machine Learning Program, Canberra, ACT 0200,
Australia; and ANU}
\icmlauthor{Arthur Gretton}{arthur.gretton@tuebingen.mpg.de}
\icmladdress{MPI for Biological Cybernetics, Spemannstr.\ 38,
72076 T\"ubingen, Germany}
\icmlauthor{Karsten Borgwardt}{borgwardt@dbs.ifi.lmu.de}
\icmladdress{LMU, Department ''Institute for Informatics'', Oettingenstr.\
67, 80538 M\"unchen, Germany}
\icmlauthor{Justin Bedo}{bedo@ieee.org} \icmladdress{NICTA,
Statistical Machine Learning Program, Canberra, ACT 0200,
Australia}
 \vskip 0.1in ]

\begin{abstract}
We introduce a framework for filtering features that employs the
Hilbert-Schmidt Independence Criterion (HSIC) as a measure of
dependence between the features and the labels. The key idea is
that good features should maximise such dependence. Feature
selection for various supervised learning problems (including
classification and regression) is unified under this framework,
and the solutions  can be approximated using a
backward-elimination algorithm. We demonstrate the usefulness of
our method on both artificial and real world datasets.
\end{abstract}

\section{Introduction}
\label{se:intro}

In supervised learning problems, we
are typically given $m$ data points $x\in\mathcal{X}$ and their
labels $y\in\mathcal{Y}$. The task is to find a functional
dependence between $x$ and $y$, $f:x\longmapsto y$, subject to
certain optimality conditions. Representative tasks include binary
classification, multi-class classification, regression and
ranking. We often want to reduce the dimension of the data
(the number of features) before the actual learning
\citep{GuyEli03}; a larger number of features  can be associated
with higher data collection cost, more difficulty in
model interpretation,  higher computational cost for the
classifier, and decreased generalisation ability. It is
therefore important to select an informative feature subset.

The problem of supervised feature selection can be cast as a
combinatorial optimisation problem. We have a full set of
features, denoted $\Scal$ (whose elements correspond to the
dimensions of the data).  We use these features to predict a
particular outcome, for instance the presence of cancer: clearly,
only a subset $\Tcal$ of features will be relevant. Suppose the
relevance of $\Tcal$ to the outcome is quantified by
$\Qcal(\Tcal)$, and is computed by restricting the data to the
dimensions in $\Tcal$. Feature
selection can then be formulated as\\[-0.5cm]
\begin{align}
\Tcal_0=\arg\max_{\Tcal\subseteq\Scal}~ \Qcal(\Tcal)\qquad \text{
subject to } \quad |\Tcal|\leq t, \vspace{-1mm} \label{eq:fs}
\end{align}\\[-0.5cm]
where $|\cdot|$ computes the cardinality of a set and $t$ upper
bounds the number of selected features. Two important aspects of
problem (\ref{eq:fs}) are the choice of the criterion
$\Qcal(\Tcal)$ and the selection algorithm.

\paragraph{Feature Selection Criterion.}
The choice of $\Qcal(\Tcal)$ should respect the underlying
supervised learning tasks --- estimate  dependence function $f$
from training data and guarantee $f$ predicts well on test data.
Therefore, good criteria should satisfy two conditions:\\[-0.5cm]
\begin{enumerate*}
  \item[\textbf{I}:] $\Qcal(\Tcal)$ is capable of
    detecting any desired (nonlinear as well as linear) functional
    dependence between the data and  labels.
  \item[\textbf{II}:] $\Qcal(\Tcal)$ is concentrated with respect to the
    underlying measure. This guarantees with high probability that the
    detected functional dependence is preserved in the test data.
\end{enumerate*}
While many feature selection criteria have been explored, few take these two conditions
explicitly into account. Examples include the leave-one-out error
bound of SVM \citep{WesMukChaetal00} and
%
%
the mutual information \citep{KolSah96}. Although the latter has
good theoretical justification, it requires density estimation,
which is problematic for high dimensional and continuous
variables. We sidestep these problems by employing a
mutual-information \emph{like} quantity --- the Hilbert Schmidt
Independence Criterion (HSIC) \citep{GreBouSmoetal05}. HSIC uses
kernels for measuring dependence and does not require density
estimation. HSIC also has good uniform convergence guarantees. As
we show in section~\ref{se:hsic}, HSIC satisfies conditions
\textbf{I} and \textbf{II}, required for $\Qcal(\Tcal)$.

\paragraph{Feature Selection Algorithm.} Finding a global optimum for
\eq{eq:fs} is in general NP-hard \citep{WesEliSchetal03}. Many
algorithms transform \eq{eq:fs} into a continuous problem by
introducing weights on the dimensions
\citep{WesMukChaetal00,WesEliSchetal03}. These methods perform
well for linearly separable problems. For nonlinear problems,
however, the optimisation usually becomes non-convex and a local
optimum does not necessarily provide good features. Greedy
approaches -- forward selection and backward elimination -- are often
used to tackle problem (\ref{eq:fs}) directly. Forward selection
tries to increase $\Qcal(\Tcal)$ as much as possible for each
inclusion of features, and backward elimination tries to achieve
this for each deletion of features~\citep{GuyWesBarVap02}.
Although forward selection is
computationally more efficient, backward elimination provides
better features in general since the features are
assessed within the context of all others.

\paragraph{BAHSIC.} In principle, HSIC can be employed using either
the forwards or backwards strategy, or a mix of strategies.
However, in this paper, we will focus on a backward elimination
algorithm. Our experiments show that backward elimination
outperforms forward selection for HSIC. Backward elimination using
HSIC (BAHSIC) is a filter method for feature selection. It selects
features independent of a particular classifier. Such decoupling
not only facilitates subsequent feature interpretation but also
speeds up the computation over wrapper and embedded methods.

Furthermore, BAHSIC is directly applicable to binary, multiclass, and
regression problems. Most other feature selection methods are only
formulated either for binary classification or regression.  The multi-class
extension of these methods is usually accomplished using a
one-versus-the-rest strategy. Still fewer methods handle classification
and regression cases at the same time. BAHSIC, on the other hand,
accommodates all these cases in a principled way: by choosing different
kernels, BAHSIC also subsumes many existing methods as special cases.
The versatility of BAHSIC originates from the generality of HSIC.
Therefore, we begin our exposition with an introduction of HSIC.

\section{Measures of Dependence}
\label{se:hsic}
We define $\Xcal$ and $\Ycal$ broadly as two domains from which we
draw samples $(x,y)$: these may be real valued, vector valued,
class labels, strings,
graphs,
and so on.
We define a (possibly nonlinear) mapping $\phi(x)\in\Fcal$ from
each $x\in \Xcal$ to a feature space $\Fcal$, such that the inner
product between the features is given by a kernel function
$k(x,x'):=\langle\phi(x),\phi(x')\rangle$: $\Fcal$ is called a
reproducing kernel Hilbert space (RKHS). Likewise, let $\Gcal$ be
a second RKHS on $\mathcal{Y}$ with kernel $l(\cdot,\cdot)$ and
feature map $\psi(y)$. We may now define a cross-covariance
operator between these feature maps, in accordance with
\citet{Bak73,FukBacJor04}: this is a linear operator
$\Ccal_{xy}:\Gcal\longmapsto\Fcal$ such that
\begin{equation}
\Ccal_{xy}=\EE_{xy}[(\phi(x)-\mu_x)\otimes(\psi(y)-\mu_y)],
\end{equation}
where $\otimes$ is the tensor product. The square of the
Hilbert-Schmidt norm of the cross-covariance operator (HSIC),
$\|\Ccal_{xy}\|^2_{\rm HS}$, is then used as our feature selection
criterion $\Qcal(\Tcal)$. \citet{GreBouSmoetal05} show that HSIC
can be expressed in terms of kernels as
\begin{align}
\label{eq:def_hsic}
&\hsic(\Fcal,\Gcal,\Pr_{xy})=\|\Ccal_{xy}\|_{\rm HS}^2\\
&=~\EE_{xx'yy'}[k(x,x')l(y,y')]+
\EE_{xx'}[k(x,x')]\EE_{yy'}[l(y,y')]\nonumber\\
&-2\EE_{xy}[\EE_{x'}[k(x,x')]\EE_{y'}[l(y,y')]],\nonumber
\end{align}
where $\textsf{E}_{xx'yy'}$ is the expectation over both
$(x,y)\sim \Pr_{xy}$ and an additional pair of variables
 $(x',y')\sim\Pr_{xy}$ drawn {\em independently} according to the same law.
Previous work used HSIC to \emph{measure} independence between two
sets of random variables~\citep{GreBouSmoetal05}. Here we use it
to \emph{select} a subset $\Tcal$ from the first full set of
random variables $\Scal$. We now describe further properties of
HSIC which support its use as a feature selection criterion.

\paragraph{Property (I)} \citet[Theorem 4]{GreBouSmoetal05} show that
whenever $\Fcal,\Gcal$ are RKHSs with universal kernels $k,l$ on
respective compact domains $\mathcal{X}$ and $\mathcal{Y}$ in the
sense of \cite{Steinwart01a}, then ${\rm
HSIC}(\Fcal,\Gcal,\Pr_{xy})=0$ if and only if $x$ and $y$ are
independent. In terms of  feature selection,
 a universal kernel such as the Gaussian RBF kernel or the
Laplace kernel permits HSIC to detect any dependence between
$\Xcal$ and $\Ycal$. HSIC is zero if and only if features and labels are
independent.

In fact, non-universal kernels can also be used for HSIC,
although they may not guarantee that all dependencies are
detected. Different kernels incorporate distinctive prior
knowledge into the dependence estimation, and they focus HSIC on
dependence of a certain type. For instance, a linear kernel requires
HSIC to seek only second order dependence. Clearly HSIC is
capable of finding and exploiting dependence of a much more
general nature by kernels on graphs, strings, or other discrete domains.

\paragraph{Property (II)}
Given a sample $Z=\{(x_1,y_1),\ldots,(x_m,y_m)\}$ of size $m$
drawn from $\Pr_{xy}$, we derive an unbiased estimate of HSIC,
\begin{align}
\label{eq:e_uhsic} &\hsic(\Fcal,\Gcal,Z) \\
&=\smallfrac{1}{m(m-3)}[\tr (\mathbf{KL})+
\smallfrac{\one^\top\Kb\one\one^\top\Lb\one}{(m-1)(m-2)}-\smallfrac{2}{m-2}\one^\top\Kb\Lb\one
], \nonumber
\end{align}
where $\Kb$ and $\Lb$ are computed as
$\Kb_{ij}=(1-\delta_{ij})k(x_i,x_j)$ and
$\Lb_{ij}=(1-\delta_{ij})l(y_i,y_j)$. Note that the diagonal
entries of $\Kb$ and $\Lb$ are set to zero. The following theorem,
a formal statement that the empirical HSIC is unbiased, is proved
in the appendix.\\[-0.5cm]
\begin{theorem}[HSIC is Unbiased]
\label{th:unbias}
Let
  $\EE_{Z}$
  denote the expectation taken over $m$ independent
  observations $(x_i,y_i)$ drawn from $\Pr_{xy}$. Then
  \begin{align}
    \hsic(\Fcal,\Gcal,\Pr_{xy})=\EE_{Z}\sbr{\hsic(\Fcal,\Gcal,Z)}.
  \end{align}
\end{theorem}
This property is by contrast with the mutual information, which
can require sophisticated bias correction strategies
\citep[e.g.][]{NemShaBia02}.

\paragraph{U-Statistics.}
The estimator in (\ref{eq:e_uhsic}) can be alternatively formulated
using U-statistics,
\begin{align}
  \hsic(\Fcal,\Gcal,Z)
  = (m)_4^{-1}\sum_{(i,j,q,r)\in\mathbf{i}_{4}^{m}}^{m}h(i,j,q,r),
  \label{eq:ustats}
\end{align}
where $(m)_n = \frac{m!}{(m-n)!}$ is the Pochhammer coefficient
and where $\mathbf{i}_{r}^{m}$ denotes the set of all $r$-tuples
drawn without replacement from $\{1,\ldots,m\}$. The kernel $h$ of
the U-statistic is defined by
\begin{align}
\frac{1}{4!}\sum_{(s,t,u,v)}^{(i,j,q,r)}\rbr{\Kb_{st}\Lb_{st}+\Kb_{st}\Lb_{uv}-2\Kb_{st}\Lb_{su}},
\label{eq:u_kernel}
\end{align}
where the sum in \eq{eq:u_kernel} represents all ordered
quadruples $(s,t,u,v)$ selected without replacement from
$(i,j,q,r)$.
%

We now show that $\hsic(\Fcal,\Gcal,Z)$ is concentrated.
Furthermore, its convergence in probability to
$\hsic(\Fcal,\Gcal,\Pr_{xy})$ occurs with rate $1/\sqrt{m}$ which
is a slight improvement over the convergence of the biased
estimator by \cite{GreBouSmoetal05}.

\begin{theorem}[HSIC is Concentrated]
\label{th:concentration}
Assume $k,l$ are bounded almost everywhere by $1$, and are
non-negative. Then for $m>1$ and all $\delta>0$, with probability
at least $1-\delta$ for all $\Pr_{xy}$
\begin{align*}
|\hsic(\Fcal,\Gcal,Z)-\hsic(\Fcal,\Gcal,\Pr_{xy})|\leq 8
 \sqrt{\log(2/\delta) / m}
\end{align*}
\end{theorem}
By virtue of \eq{eq:ustats} we see immediately that $\hsic$ is a
U-statistic of order 4, where each term is bounded in $[-2, 2]$.
Applying Hoeffing's bound as in \cite{GreBouSmoetal05} proves the
result.

These two theorems imply the empirical HSIC closely
reflects its population counterpart. This means the same features
should consistently be selected to achieve high dependence if the
data are repeatedly drawn from the same distribution.


\paragraph{Asymptotic Normality.}
It follows from \cite{Serfling80} that under the assumptions
$\EE(h^2)<\infty$ and that the data and labels are not
independent, the empirical HSIC converges in distribution to a
Gaussian random variable with mean $\hsic(\Fcal,\Gcal,\Pr_{xy})$
and variance
\begin{align}
  \label{eq:varHSIC}
  \sigma_\hsic^{2} & =\smallfrac{16}{m}\left(R-\mathrm{HSIC}^{2}\right),
  \text{ where } \\
  \nonumber
  R & =
  \smallfrac{1}{m}\!\sum_{i=1}^{m}\Bigl( (m-1)_3^{-1}\!\!\sum_{(j,q,r)\in\mathbf{i}_{3}^{m}\setminus\{
i\}}\!\!\!h(i,j,q,r)\Bigr)^{2},
\end{align}
and $\mathbf{i}_{r}^{m}\setminus\{i\}$ denotes the set of all
$r$-tuples drawn without replacement from
$\{1,\ldots,m\}\setminus\{i\}$. The asymptotic normality
allows us to formulate statistics for a significance test.
This is useful because it may provide an assessment of the
dependence between the selected features and the labels.

\paragraph{Simple Computation.}
Note that $\hsic(\Fcal,\Gcal,Z)$ is simple to compute, since only
the kernel matrices $\Kb$ and $\Lb$ are needed, and no density
estimation is involved. For feature selection, $\Lb$ is fixed
through the whole process. It can be precomputed and stored for
speedup if needed. Note also that ${\rm HSIC}(\Fcal,\Gcal,Z)$ does
\emph{not} need any explicit regularisation parameter. This is
encapsulated in the choice of the kernels.

\section{Feature Selection via HSIC }
\label{se:bahsic}
Having defined our feature selection criterion, we now describe an
algorithm that conducts feature selection on the basis of this
dependence measure. Using HSIC, we can perform both backward
(BAHSIC) and forward (FOHSIC) selection of the features. In
particular, when we use a linear kernel on the data (there is no
such requirement for the labels), forward selection and backward
selection are equivalent: the objective function decomposes into
individual coordinates, and thus feature selection can be done
without recursion in one go. Although forward selection is
computationally more efficient, backward elimination in general
yields better features, since the quality of the features is
assessed within the context of all other features. Hence we
present the backward elimination version of our algorithm here
(a forward greedy selection version can be derived similarly).

BAHSIC appends the features from $\mathcal{S}$ to the end of a
list $\mathcal{S}^{\dagger}$ so that the elements towards the end
of $\mathcal{S}^{\dagger}$ have higher relevance to the learning
task. The feature selection problem in (\ref{eq:fs}) can be solved
by simply taking the last $t$ elements from
$\mathcal{S}^{\dagger}$. Our algorithm produces
$\mathcal{S}^{\dagger}$  recursively, eliminating the least
relevant features from $\mathcal{S}$ and adding them to the end of
$\mathcal{S}^{\dagger}$ at each iteration. For convenience, we
also denote HSIC as $\hsic(\sigma,\mathcal{S})$, where
$\mathcal{S}$ are the features used in computing the data kernel
matrix $\mathbf{K}$, and $\sigma$ is the parameter for the data
kernel (for instance, this might be the size of a Gaussian kernel
$k(x,x')=\exp(-\sigma\nbr{x-x'}^2)$ ).
\begin{algorithm}[ht]
\caption{BAHSIC}
\textbf{Input}: The full set of features $\Scal$\\
\textbf{Output}: An ordered set of features
$\mathcal{S}^{\dagger}$\\[-0.4cm] \label{alg:abopt}
\begin{algorithmic}[1]
    \STATE $\Scal^{\dagger}\leftarrow\varnothing$
    \REPEAT
        \STATE $\sigma\leftarrow\Xi$
        \STATE $\Ical \leftarrow \arg\max_{\Ical}~\sum_{j\in \Ical}\hsic(\sigma,\Scal\setminus\{j\} ),~~\Ical\subset\Scal$
        \STATE $\Scal\leftarrow\Scal\setminus\Ical$
        \STATE $\Scal^{\dagger}\leftarrow\Scal^{\dagger}\cup\Ical$
    \UNTIL{$\Scal=\varnothing$}
\end{algorithmic}
\end{algorithm}

Step 3 of the algorithm denotes a policy for adapting the kernel
parameters, e.g. by optimising over the
possible parameter choices. In our experiments, we
typically normalize each feature separately to zero mean and unit
variance, and adapt the parameter for a Gaussian kernel
by setting $\sigma$ to $1/(2d)$, where $d=|\Scal|-1$.
 If we have prior knowledge about the type of nonlinearity, we
can use a kernel with fixed parameters for BAHSIC. In this case,
step 3 can be omitted.

Step 4 of the algorithm is concerned with the selection of a set
$\Ical$ of features to eliminate. While one could choose a single
element of $\Scal$, this would be inefficient when there are a
large number of irrelevant features. On the other hand, removing
too many features at once risks the loss of relevant features. In
our experiments, we found a good compromise between speed and
feature quality was to remove 10\% of the current features at each
iteration.

\section{Connections to Other Approaches}
\label{se:connection}

We now explore connections to other feature selectors. For binary
classification, an alternative criterion for selecting features is
to check whether the distributions $\Pr(x|y=1)$ and $\Pr(x|y=-1)$
differ. For this purpose one could use Maximum Mean Discrepancy
(MMD)~\citep{BorGreRasKriSchSmo06}.  Likewise, one could use
Kernel Target Alignment (KTA) \citep{CriKanEliSha03} to test
directly whether there exists any correlation between data and
labels. KTA has been used for feature selection.  Formally it is
defined as $\tr \Kb \Lb / \|\Kb\| \|\Lb\|$. For computational
convenience the normalisation is often omitted in practise
\citep{NeuSchSte05}, which leaves us with $\tr \Kb \Lb$.  We
discuss this unnormalised variant below.

Let us consider the output kernel $l(y,y') = \rho(y) \rho(y')$,
where $\rho(1) = m_+^{-1}$ and $\rho(-1) = -m_-^{-1}$, and $m_+$
and $m_-$ are the numbers of positive and negative samples,
respectively. With this kernel choice, we show that MMD and KTA
are closely related to HSIC. The following theorem is proved in
the appendix.

\begin{theorem}[Connection to MMD and KTA]
  \label{th:mmdkta}
  Assume the kernel $k(x,x')$ for the data is bounded and the kernel for
  the labels is $l(y,y')=\rho(y)\rho(y')$. Then
  \begin{align*}
  \left|\hsic-(m-1)^{-2}{\rm MMD}\right| & =O(m^{-1})  \\
  \left|\hsic-(m-1)^{-2}{\rm KTA}\right| & =O(m^{-1}).
  \end{align*}
\end{theorem}
This means selecting features that maximise HSIC also maximises
MMD and KTA. Note that in general (multiclass, regression, or
generic binary classification) this connection does not hold.

\section{Variants of BAHSIC}

New variants can be readily derived from BAHSIC by combining the
two building blocks of BAHSIC: a kernel on the data and another
one on the labels. Here we provide three examples using a Gaussian
kernel on the data, while varying the kernel on the labels. This
provides us with feature selectors for three problems:

\paragraph{Binary classification} (BIN) We set $m_+^{-1}$ as the label
for positive class members, and $m_-^{-1}$ for negative class members. We then apply a
linear kernel.

\paragraph{Multiclass classification} (MUL) We apply a linear kernel
on the labels using the label vectors below, as described for a
3-class example.  Here $m_i$ is the number of samples in class $i$
and $\one_{m_i}$ denotes a vector of all ones with length $m_i$.
\begin{equation}
\Yb=\left(
\begin{matrix} \frac{\one_{m_1}}{m_{1}} & \frac{\one_{m_1}}{m_{2}-m} & \frac{\one_{m_1}}{m_{3}-m}\cr \frac{\one_{m_2}}{m_{1}-m} & \frac{\one_{m_2}}{m_{2}} & \frac{\one_{m_2}}{m_{3}-m}
\cr \frac{\one_{m_3}}{m_{1}-m} & \frac{\one_{m_3}}{m_{2}-m} &
\frac{\one_{m_3}}{m_{3}}
\end{matrix}
\right)_{m\times 3}.\label{eq:mc}
\end{equation}

\paragraph{Regression} (REG) A Gaussian RBF kernel
is also used on the labels. For convenience the kernel width
$\sigma$ is fixed as the median distance between points in the
sample \citep{SchSmo02}.

For the above variants a further speedup of BAHSIC is possible by
updating the entries of the kernel matrix incrementally, since we
are using an RBF kernel. We use the fact that
$\|x-x'\|^2=\sum_j\|x_j-x_j'\|^2$. Hence $\|x-x'\|^2$ needs to be
computed only once. Subsequent updates are effected by subtracting
$\|x_j-x_j'\|^2$ (subscript here indices dimension).

We will use BIN, MUL and REG as the particular instances of BAHSIC
in our  experiments. We will refer to them commonly as BAHSIC
since the exact meaning will be clear depending on
the datasets encountered. Furthermore, we also instantiate FOHSIC
using the same kernels as BIN, MUL and REG, and we adopt the same
convention when we refer to it in our experiments.

\vspace{-2mm}
\section{Experimental Results}
\label{se:experiments}
\vspace{-2mm}

We conducted three sets of experiments. The characteristics of the
 datasets and the aims of the experiments are: (\emph{i})
artificial datasets illustrating the properties of BAHSIC;
(\emph{ii}) real datasets that compare BAHSIC with other methods;
and (\emph{iii}) a brain computer interface dataset showing that
BAHSIC selects meaningful features.

\begin{figure}[bt]
\includegraphics[width=0.45\columnwidth]{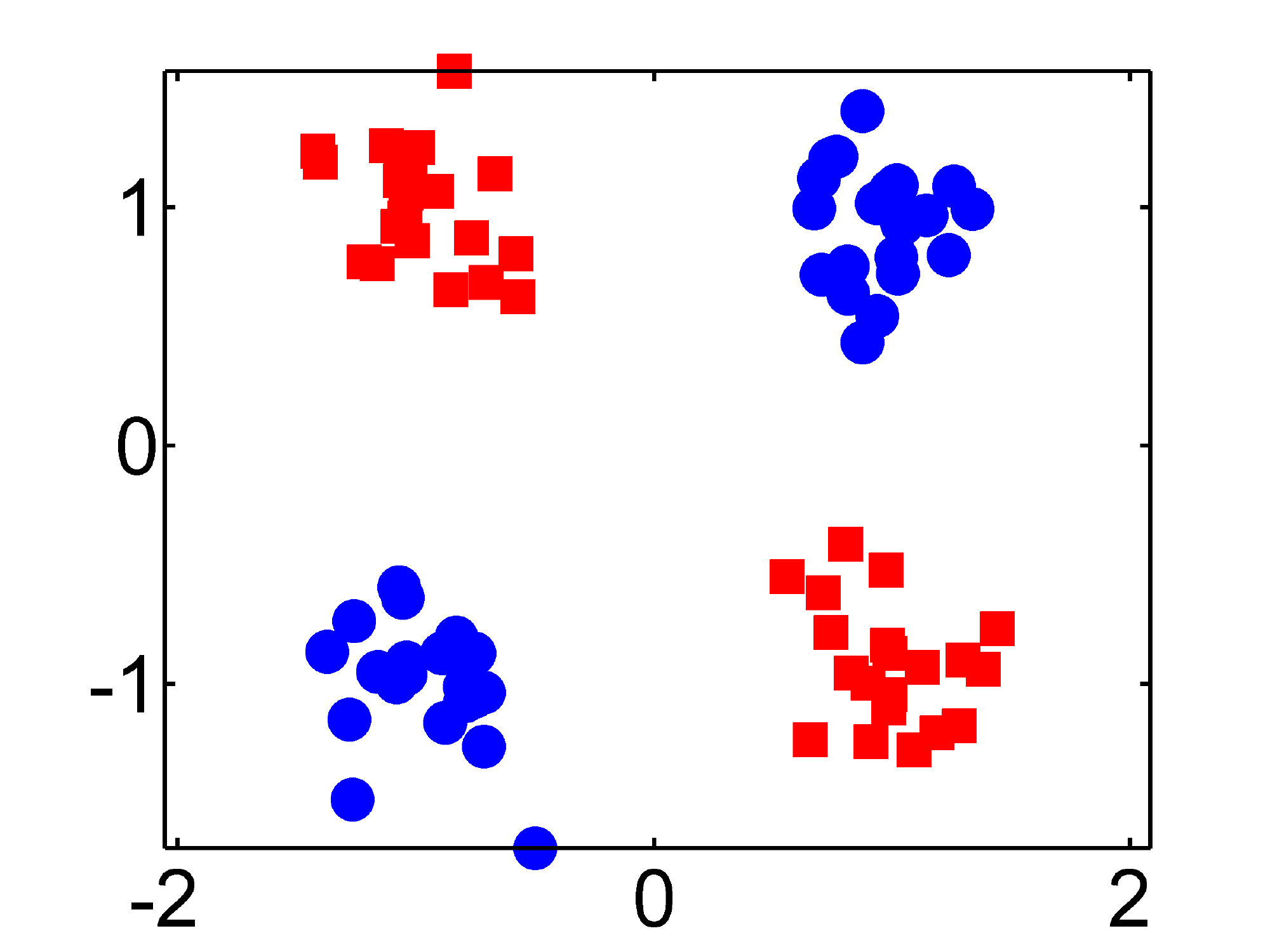}
\hspace{0.05\columnwidth}
\includegraphics[width=0.45\columnwidth]{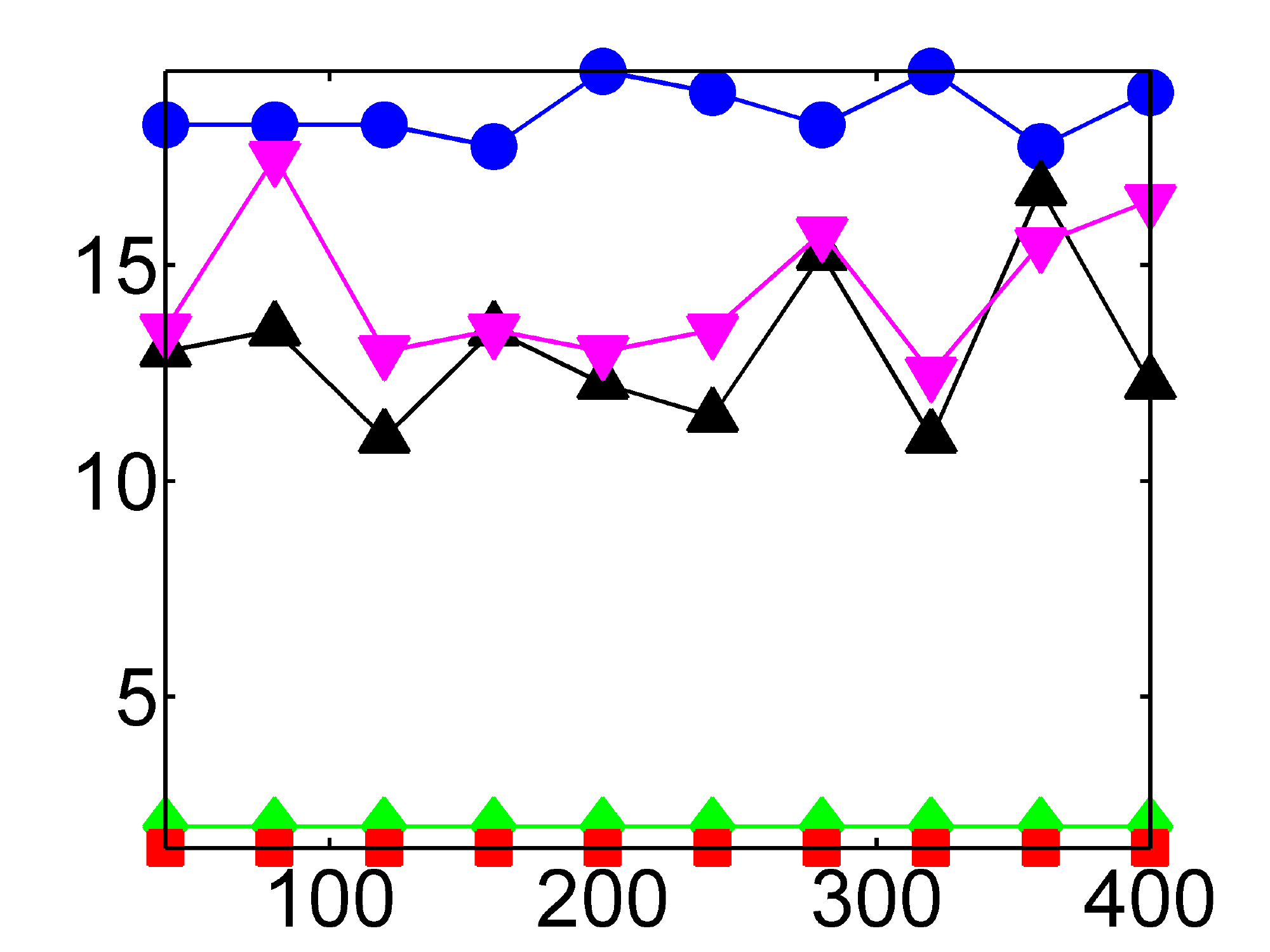} \\
\includegraphics[width=0.45\columnwidth]{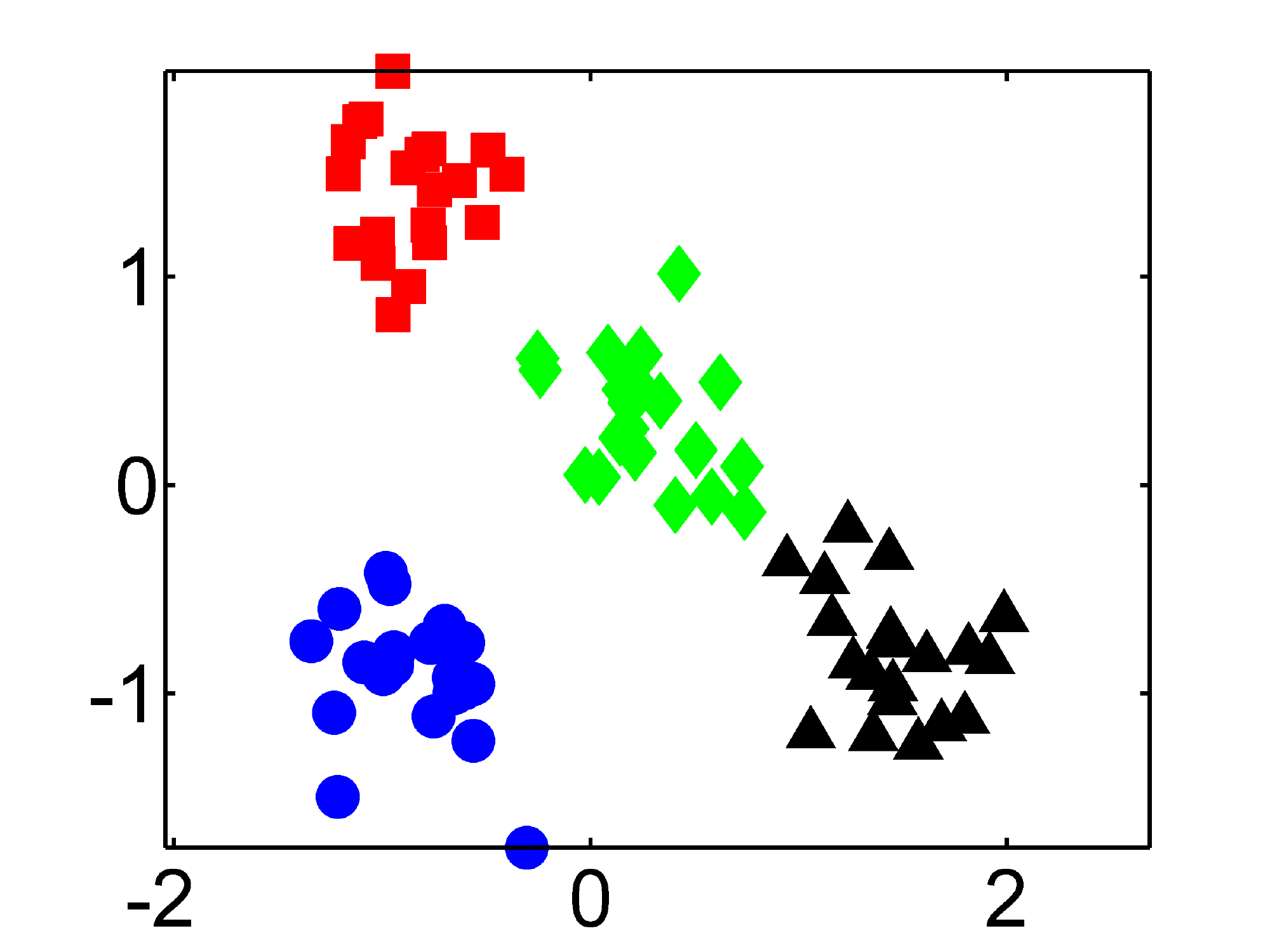}
\hspace{0.05\columnwidth}
\includegraphics[width=0.45\columnwidth]{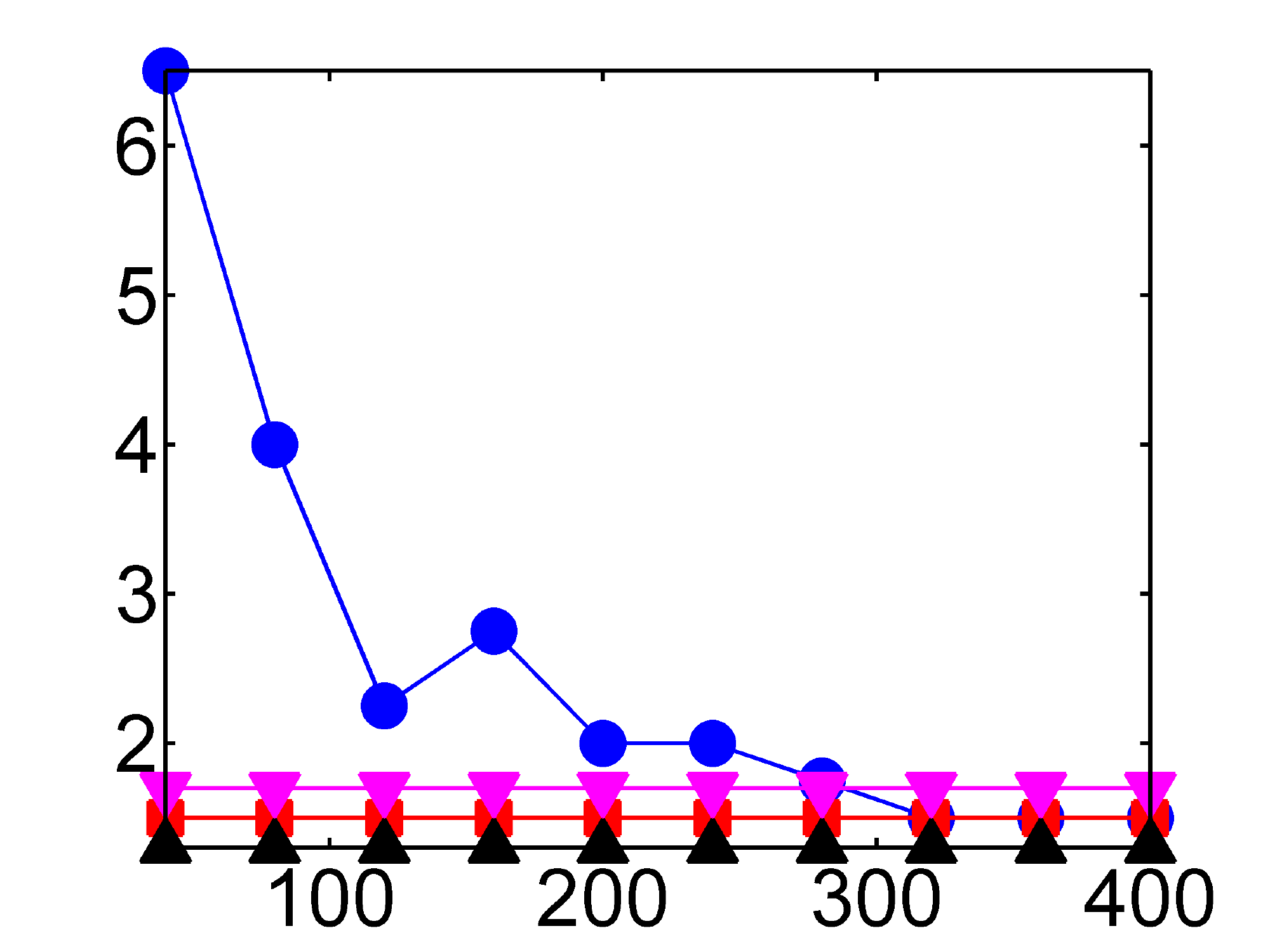} \\
\includegraphics[width=0.45\columnwidth]{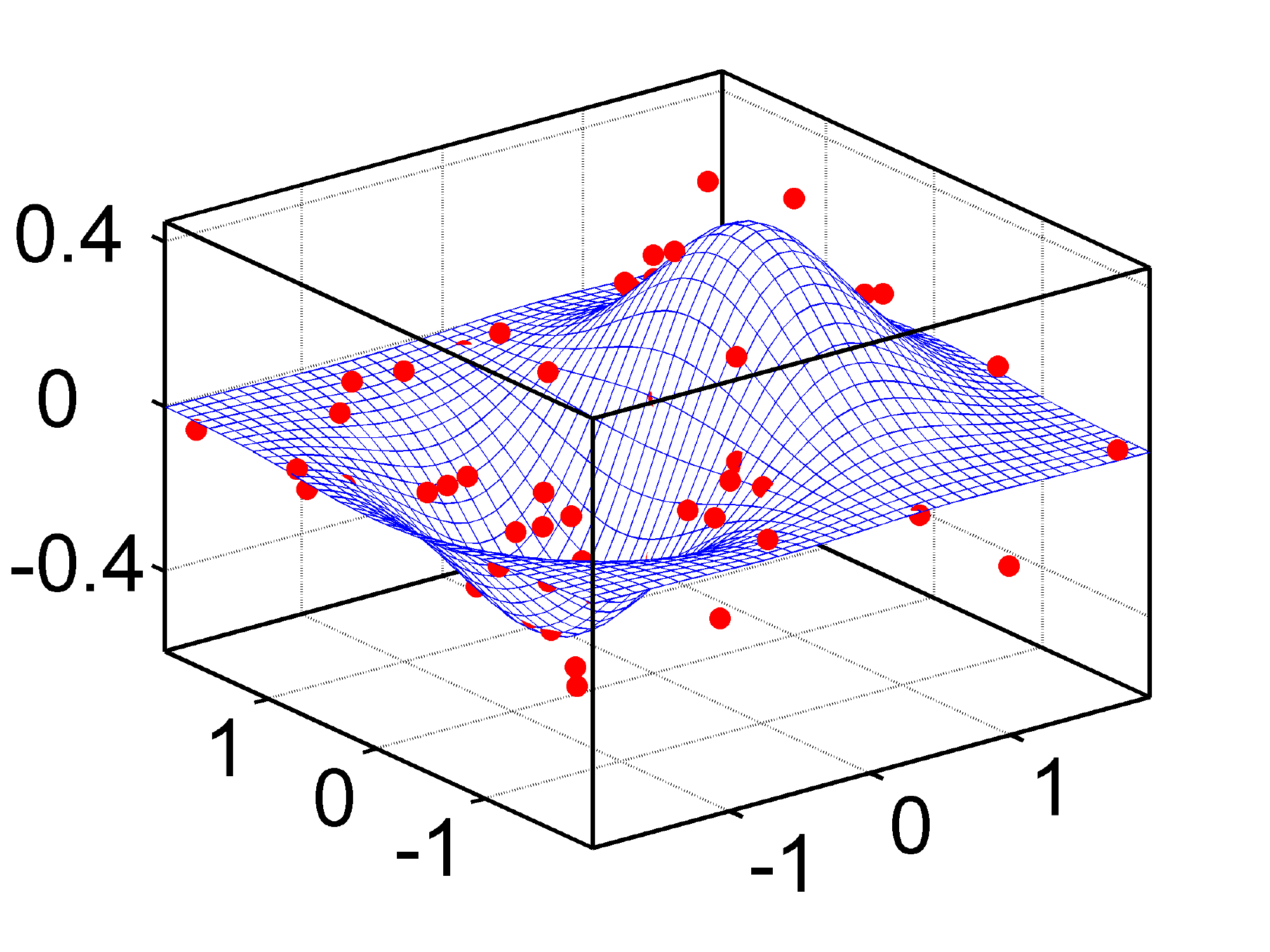}
\hspace{0.05\columnwidth}
\includegraphics[width=0.45\columnwidth]{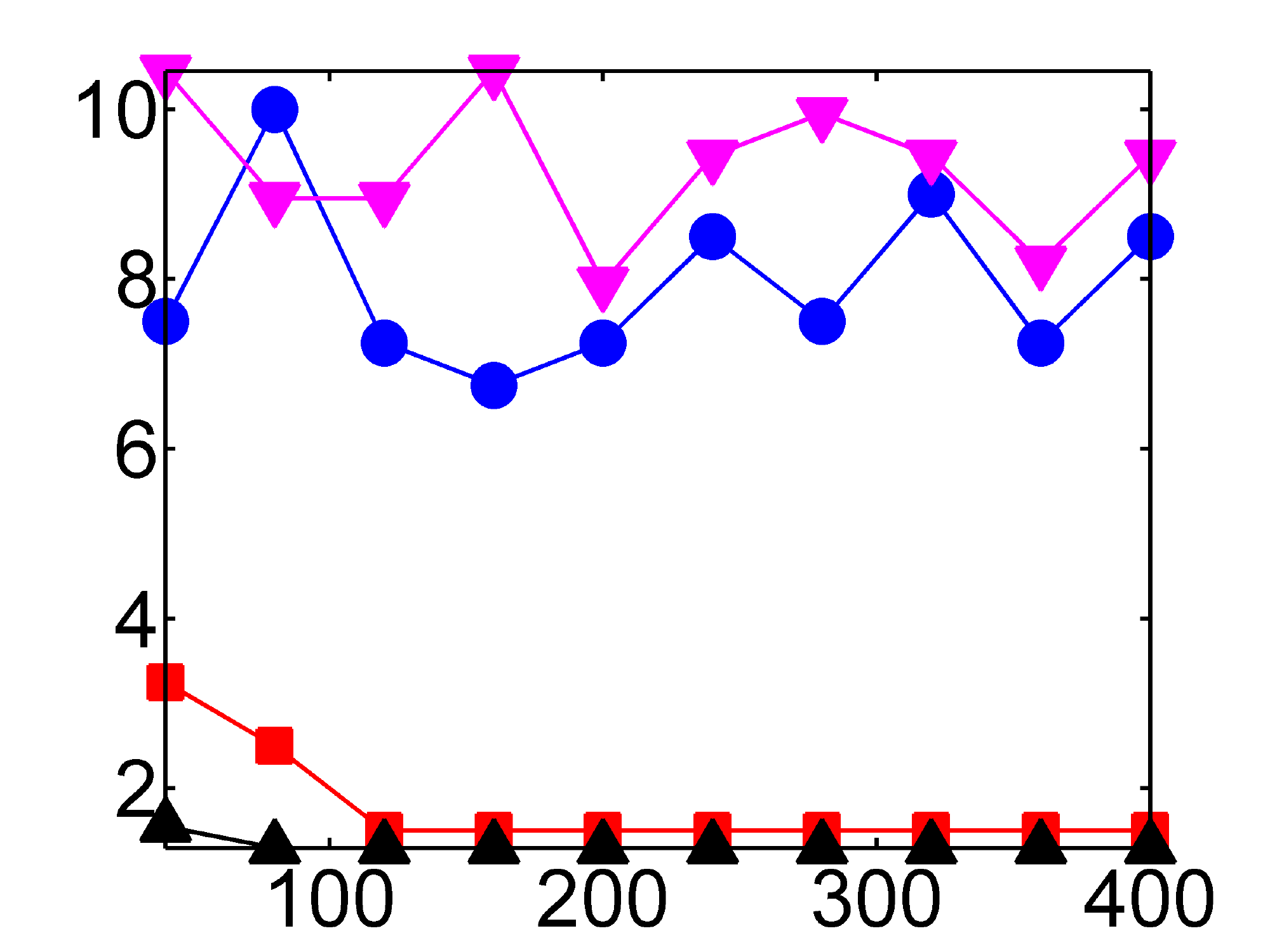}
\caption{Artificial datasets and the performance of different
methods when varying the number of observations.
{\bfseries Left column, top to bottom:} Binary, multiclass, and
regression data. Different classes are encoded with different colours.
{\bfseries Right column:} Median rank (y-axis) of the two relevant
features as a function of sample size (x-axis) for the
corresponding datasets in the left column. (Blue circle: Pearson's
correlation; Green triangle: RELIEF; Magenta downward triangle:
mutual information; Black triangle: FOHSIC; Red square: BAHSIC.)}
\label{fg:artificial}
\end{figure}

\vspace{-2mm}
\subsection{Artificial datasets}
\label{ex:artificial}
\vspace{-2mm}

We constructed 3 artificial datasets, as illustrated in
Figure~\ref{fg:artificial}, to illustrate the difference between BAHSIC
variants with linear and nonlinear kernels. Each dataset has 22
dimensions --- only the first two dimensions are related to the
prediction task and the rest are just Gaussian noise. These
datasets are (\textit{i}) \textbf{Binary XOR data}: samples
belonging to the same class have multimodal distributions;
(\textit{ii}) \textbf{Multiclass data}: there are 4 classes but 3
of them are collinear; (\textit{iii}) \textbf{Nonlinear regression
data}: labels are related to the first two dimension of the data by
$y=x_1\exp(-x_1^2-x_2^2)+\epsilon$, where $\epsilon$ denotes
additive Gaussian noise. We compare BAHSIC to FOHSIC, Pearson's
correlation, mutual information \citep{ZafHut02}, and RELIEF (RELIEF works
only for binary problems). We aim to show that when nonlinear
dependencies exist in the data, BAHSIC with nonlinear kernels is
very competent in finding them.

We instantiate the artificial datasets over a range of sample
sizes (from 40 to 400), and  plot the median rank, produced by
various methods, for the first two dimensions of the data. All
numbers in Figure~\ref{fg:artificial} are averaged over 10 runs.
In all cases, BAHSIC shows good performance. More specifically, we
observe:

\paragraph{Binary XOR} Both BAHSIC and RELIEF correctly select the first
two dimensions of the data even for small sample sizes; while FOHSIC,
Pearson's correlation, and mutual information fail. This
is because the latter three evaluate the goodness of each feature
independently. Hence they are unable to capture nonlinear
interaction between features.

\paragraph{Multiclass Data} BAHSIC, FOHSIC and mutual information
select the correct features irrespective of the size of the
sample. Pearson's correlation only works for large sample size.
The collinearity of 3 classes provides linear correlation between
the data and the labels, but due to the interference of the fourth
class such correlation is picked up by Pearson's correlation only
for a large sample size.

\paragraph{Nonlinear Regression Data} The performance of Pearson's correlation
and mutual information is slightly better than random. BAHSIC and
FOHSIC quickly converge to the correct answer as the sample size
increases.

In fact, we observe that as the sample size increases, BAHSIC is
able to rank the relevant features (the first two dimensions)
almost correctly in the first iteration (results not shown). While
this does not prove BAHSIC with nonlinear kernels is always better
than that with a linear kernel, it illustrates the competence of
BAHSIC in detecting nonlinear features. This is obviously useful
in a real-world situations. The second advantage of BAHSIC is that
it is readily applicable to both classification and regression
problems, by simply choosing a different kernel on the labels.

\begin{table*}[tb]
\scriptsize \caption{Classification error (\%) or percentage of
variance \emph{not}-explained (\%). The best result, and those
results not significantly worse than it, are highlighted in bold
(one-sided Welch t-test with 95\% confidence level).
100.0$\pm$0.0$^\ast$: program is not finished in a week or
crashed. -: not applicable. }\centering \label{tb:real}

\begin{tabular}{|c|*{8}{r@{$\pm$}l|}}
\hline \textbf{Data} & \twoco{\textbf{BAHSIC}} &
\twoco{\textbf{FOHSIC}} & \twoco{\textbf{PC}} &
\twoco{\textbf{MI}} & \twoco{\textbf{RFE}} &
\twoco{\textbf{RELIEF}} &
\twoco{$\L_0$}  & \twoco{\textbf{R2W2}} \\
\hline
covertype   &   \textbf{26.3}&\textbf{1.5}  &   37.9&1.7    &   40.3&1.3    &   \textbf{26.7}&\textbf{1.1}  &   33.0&1.9    &   42.7&0.7    &   43.4&0.7    &   44.2&1.7    \\
ionosphere  &   \textbf{12.3}&\textbf{1.7}  &   \textbf{12.8}&\textbf{1.6}  &   \textbf{12.3}&\textbf{1.5}  &   \textbf{13.1}&\textbf{1.7}  &   20.2&2.2    &   \textbf{11.7}&\textbf{2.0}  &   35.9&0.4    &   13.7&2.7    \\
sonar       &   27.9&3.1    &   25.0&2.3    &   25.5&2.4    &   26.9&1.9    &   \textbf{21.6}&\textbf{3.4}  &   24.0&2.4    &   36.5&3.3    &   32.3&1.8    \\
heart       &   \textbf{14.8}&\textbf{2.4}  &   \textbf{14.4}&\textbf{2.4}  &   16.7&2.4    &   \textbf{15.2}&\textbf{2.5}  &   21.9&3.0    &   21.9&3.4    &   30.7&2.8    &   19.3&2.6    \\
breastcancer&   3.8&0.4 &   3.8&0.4 &   4.0&0.4 &   3.5&0.5 &   \textbf{3.4}&\textbf{0.6}   &   \textbf{3.1}&\textbf{0.3}   &   32.7&2.3    &   \textbf{3.4}&\textbf{0.4}   \\
australian  &   \textbf{14.3}&\textbf{1.3}  &   \textbf{14.3}&\textbf{1.3}  &   \textbf{14.5}&\textbf{1.3}  &   \textbf{14.5}&\textbf{1.3}  &   \textbf{14.8}&\textbf{1.2}  &   \textbf{14.5}&\textbf{1.3}  &   35.9&1.0    &   \textbf{14.5}&\textbf{1.3}  \\
splice      &   22.6&1.1    &   22.6&1.1    &   22.8&0.9    &   21.9&1.0    &   \textbf{20.7}&\textbf{1.0}  &   22.3&1.0    &   45.2&1.2    &   24.0&1.0    \\
svmguide3   &   \textbf{20.8}&\textbf{0.6}  &   \textbf{20.9}&\textbf{0.6}  &   21.2&0.6    &   \textbf{20.4}&\textbf{0.7}  &   21.0&0.7    &   21.6&0.4    &   23.3&0.3    &   23.9&0.2    \\
adult       &   24.8&0.2    &   24.4&0.6    &   \textbf{18.3}&\textbf{1.1}  &   21.6&1.1    &   21.3&0.9    &   24.4&0.2    &   24.7&0.1    &   100.0&0.0$^\ast$   \\
cleveland   &   \textbf{19.0}&\textbf{2.1}  &   \textbf{20.5}&\textbf{1.9}  &   21.9&1.7    &   \textbf{19.5}&\textbf{2.2}  &   20.9&2.1    &   22.4&2.5    &   25.2&0.6    &   21.5&1.3    \\
derm        &   \textbf{0.3}&\textbf{0.3}   &   \textbf{0.3}&\textbf{0.3}   &   \textbf{0.3}&\textbf{0.3}   &   \textbf{0.3}&\textbf{0.3}   &   \textbf{0.3}&\textbf{0.3}   &   \textbf{0.3}&\textbf{0.3}   &   24.3&2.6    &   \textbf{0.3}&\textbf{0.3}   \\
hepatitis   &   \textbf{13.8}&\textbf{3.5}  &   \textbf{15.0}&\textbf{2.5}  &   \textbf{15.0}&\textbf{4.1}  &   \textbf{15.0}&\textbf{4.1}  &   \textbf{15.0}&\textbf{2.5}  &   17.5&2.0    &   16.3&1.9    &   17.5&2.0    \\
musk        &   29.9&2.5    &   29.6&1.8    &   \textbf{26.9}&\textbf{2.0}  &   31.9&2.0    &   34.7&2.5    &   \textbf{27.7}&\textbf{1.6}  &   42.6&2.2    &   36.4&2.4    \\
optdigits   &   \textbf{0.5}&\textbf{0.2}   &   \textbf{0.5}&\textbf{0.2}   &   \textbf{0.5}&\textbf{0.2}   &   3.4&0.6 &   3.0&1.6 &   0.9&0.3 &   12.5&1.7    &   0.8&0.3 \\
specft      &   \textbf{20.0}&\textbf{2.8}  &   \textbf{20.0}&\textbf{2.8}  &   \textbf{18.8}&\textbf{3.4}  &   \textbf{18.8}&\textbf{3.4}  &   37.5&6.7    &   26.3&3.5    &   36.3&4.4    &   31.3&3.4    \\
wdbc        &   \textbf{5.3}&\textbf{0.6}   &   \textbf{5.3}&\textbf{0.6}   &   \textbf{5.3}&\textbf{0.7}   &   6.7&0.5 &   7.7&1.8 &   7.2&1.0 &   16.7&2.7    &   6.8&1.2 \\
wine        &   \textbf{1.7}&\textbf{1.1}   &   \textbf{1.7}&\textbf{1.1}   &   \textbf{1.7}&\textbf{1.1}   &   \textbf{1.7}&\textbf{1.1}   &   3.4&1.4 &   4.2&1.9 &   25.1&7.2    &   \textbf{1.7}&\textbf{1.1}   \\
german      &   29.2&1.9    &   29.2&1.8    &   26.2&1.5    &   26.2&1.7    &   27.2&2.4    &   33.2&1.1    &   32.0&0.0    &   \textbf{24.8}&\textbf{1.4}  \\
gisette     &   \textbf{12.4}&\textbf{1.0}  &   \textbf{13.0}&\textbf{0.9}  &   16.0&0.7    &   50.0&0.0    &   42.8&1.3    &   16.7&0.6    &   42.7&0.7    &   100.0&0.0$^\ast$   \\
arcene      &   \textbf{22.0}&\textbf{5.1}  &   \textbf{19.0}&\textbf{3.1}  &   31.0&3.5    &   45.0&2.7    &   34.0&4.5    &   30.0&3.9    &   46.0&6.2    &   32.0&5.5    \\
madelon     &   \textbf{37.9}&\textbf{0.8}  &   \textbf{38.0}&\textbf{0.7}  &   38.4&0.6    &   51.6&1.0    &   41.5&0.8    &   38.6&0.7    &   51.3&1.1    &   100.0&0.0$^\ast$   \\
\hline $\ell_2$&   \twoco{\textbf{11.2}}    &   \twoco{14.8}    &
\twoco{19.7} &   \twoco{48.6}    &   \twoco{42.2}    &
\twoco{25.9}    &
\twoco{85.0}    &   \twoco{138.3}   \\
\hline \hline
satimage    & \textbf{15.8}&\textbf{1.0} & 17.9&0.8 & 52.6&1.7 & 22.7&0.9 & 18.7&1.3 & \twoco{-} & 22.1&1.8 & \twoco{-} \\
segment     & 28.6&1.3 & 33.9&0.9 & \textbf{22.9}&\textbf{0.5} & 27.1&1.3 & 24.5&0.8 & \twoco{-} & 68.7&7.1 & \twoco{-} \\
vehicle     & \textbf{36.4}&\textbf{1.5} & 48.7&2.2 & 42.8&1.4 &45.8&2.5 & \textbf{35.7}&\textbf{1.3} & \twoco{-} & 40.7&1.4 & \twoco{-} \\
svmguide2   & \textbf{22.8}&\textbf{2.7} & \textbf{22.2}&\textbf{2.8} & 26.4&2.5 & 27.4&1.6 & 35.6&1.3 & \twoco{-} & 34.5&1.7 & \twoco{-} \\
vowel       & \textbf{44.7}&\textbf{2.0} & \textbf{44.7}&\textbf{2.0} & 48.1&2.0 & \textbf{45.4}&\textbf{2.2} & 51.9&2.0 & \twoco{-} & 85.6&1.0 & \twoco{-} \\
usps        & \textbf{43.4}&\textbf{1.3} & \textbf{43.4}&\textbf{1.3} & 73.7&2.2 & 67.8&1.8 & 55.8&2.6 & \twoco{-} & 67.0&2.2 & \twoco{-} \\
 \hline \hline
housing     & \textbf{18.5}&\textbf{2.6} & \textbf{18.9}&\textbf{3.6} & 25.3&2.5 & \textbf{18.9}&\textbf{2.7} & \twoco{-}&\twoco{-}&\twoco{-}&\twoco{-}\\
bodyfat     & \textbf{3.5}&\textbf{2.5} & \textbf{3.5}&\textbf{2.5} & \textbf{3.4}&\textbf{2.5} & \textbf{3.4}&\textbf{2.5} & \twoco{-}&\twoco{-}&\twoco{-}&\twoco{-}\\
abalone     & \textbf{55.1}&\textbf{2.7} & \textbf{55.9}&\textbf{2.9} & \textbf{54.2}&\textbf{3.3} & 56.5&2.6 & \twoco{-}&\twoco{-}&\twoco{-}&\twoco{-}\\
\hline
\end{tabular}
\end{table*}

\vspace{-2mm}
\subsection{Real world datasets}
\vspace{-2mm}
\label{subse:realworld}
\paragraph{Algorithms}

In this experiment, we show that the performance of BAHSIC can be
comparable to other state-of-the-art feature selectors, namely SVM
Recursive Feature Elimination (RFE)~\citep{GuyWesBarVap02},
RELIEF~\citep{KirRen92}, $\L_0$-norm SVM
($\L_0$)~\citep{WesEliSchetal03}, and R2W2~\citep{WesMukChaetal00}.
We used the implementation of these algorithms as given in the
Spider machine learning toolbox, since those were the only
publicly available
implementations.\footnote{\url{http://www.kyb.tuebingen.mpg.de/bs/people/spider}}
Furthermore, we also include filter methods, namely FOHSIC,
Pearson's correlation (PC), and mutual information (MI), in our
comparisons.

\paragraph{Datasets}

We used various real world datasets taken from
the UCI repository,%
\footnote{\url{http://www.ics.uci.edu/~mlearn/MLSummary.html}}
the Statlib repository,%
\footnote{\url{http://lib.stat.cmu.edu/datasets/}}
the LibSVM website,%
\footnote{\url{http://www.csie.ntu.edu.tw/~cjlin/libsvmtools/datasets/}}
and the NIPS feature selection challenge%
\footnote{\url{http://clopinet.com/isabelle/Projects/NIPS2003/}}
for comparison. Due to scalability issues in Spider, we produced a
balanced random sample of size less than 2000 for datasets with
more than 2000 samples.

\paragraph{Experimental Protocol}

We report the performance of an SVM using a Gaussian kernel on a
feature subset of size 5 and 10-fold cross-validation. These 5
features were selected per fold using different methods. Since we
are comparing the selected features, we used the same SVM for all
methods: a Gaussian kernel with $\sigma$ set as the median
distance between points in the sample~\citep{SchSmo02} and
regularization parameter $C=100$. On classification datasets, we
measured the performance using the error rate, and on regression
datasets we used the percentage of variance \emph{not}-explained
(also known as $1-r^2$). The results for binary datasets are
summarized in the first part of Table~\ref{tb:real}. Those for
multiclass and regression datasets are reported respectively in
the second and the third parts of Table~\ref{tb:real}.

To provide a concise summary of the performance of various methods
on binary datasets, we measured how the methods compare with
the best performing one in each dataset in Table \ref{tb:real}.
We recorded the best
absolute performance of \emph{all} feature selectors as the
baseline, and computed the distance of each algorithm to the best
possible result. In this context it makes sense to penalize
catastrophic failures more than small deviations. In other words,
we would like to have a method which is at least almost always
very close to the best performing one. Taking the $\ell_2$
distance achieves  this effect, by penalizing larger
differences more heavily. It is also our goal to choose an
algorithm that performs homogeneously well across all datasets.
The $\ell_2$ distance scores are listed for the binary datasets in
Table \ref{tb:real}. In general, the smaller the $\ell_2$
distance, the better the method. In this respect, BAHSIC and
FOHSIC have the best performance. We did not produce the $\ell_2$
distance for multiclass and regression datasets, since
the limited number of such datasets did
not allow us to draw statistically significant conclusions.

\begin{table}[b!]
\setlength{\tabcolsep}{1pt} \scriptsize \caption{Classification
errors (\%) on BCI data after selecting a frequency range.}
\label{tb:bci} \centering
\begin{tabular}{|c|*{6}{r@{$\pm$}l|}}
\hline \textbf{Subject} & \twoco{\textbf{aa}} &
\twoco{\textbf{al}} & \twoco{\textbf{av}} & \twoco{\textbf{aw}} &
\twoco{\textbf{ay}}\\
\hline
\textbf{CSP}  & 17.5&2.5 & 3.1&1.2 & 32.1&2.5 & 7.3&2.7 & \textbf{6.0}&\textbf{1.6}  \\
\textbf{CSSP} & 14.9&2.9 & 2.4&1.3 & 33.0&2.7 & \textbf{5.4}&\textbf{1.9} & \textit{6.2}&\textit{1.5} \\
\textbf{CSSSP}& \textbf{12.2}&\textbf{2.1} & \textit{2.2}&\textit{0.9} & \textit{31.8}&\textit{2.8} & 6.3&1.8 & 12.7&2.0 \\
\textbf{BAHSIC} & \textit{13.7}&\textit{4.3} & \textbf{1.9}&\textbf{1.3} & \textbf{30.5}&\textbf{3.3} & \textit{6.1}&\textit{3.8} & 9.0&6.0  \\
\hline
\end{tabular}
\end{table}

 \begin{figure*}
\includegraphics[width=0.4\columnwidth]{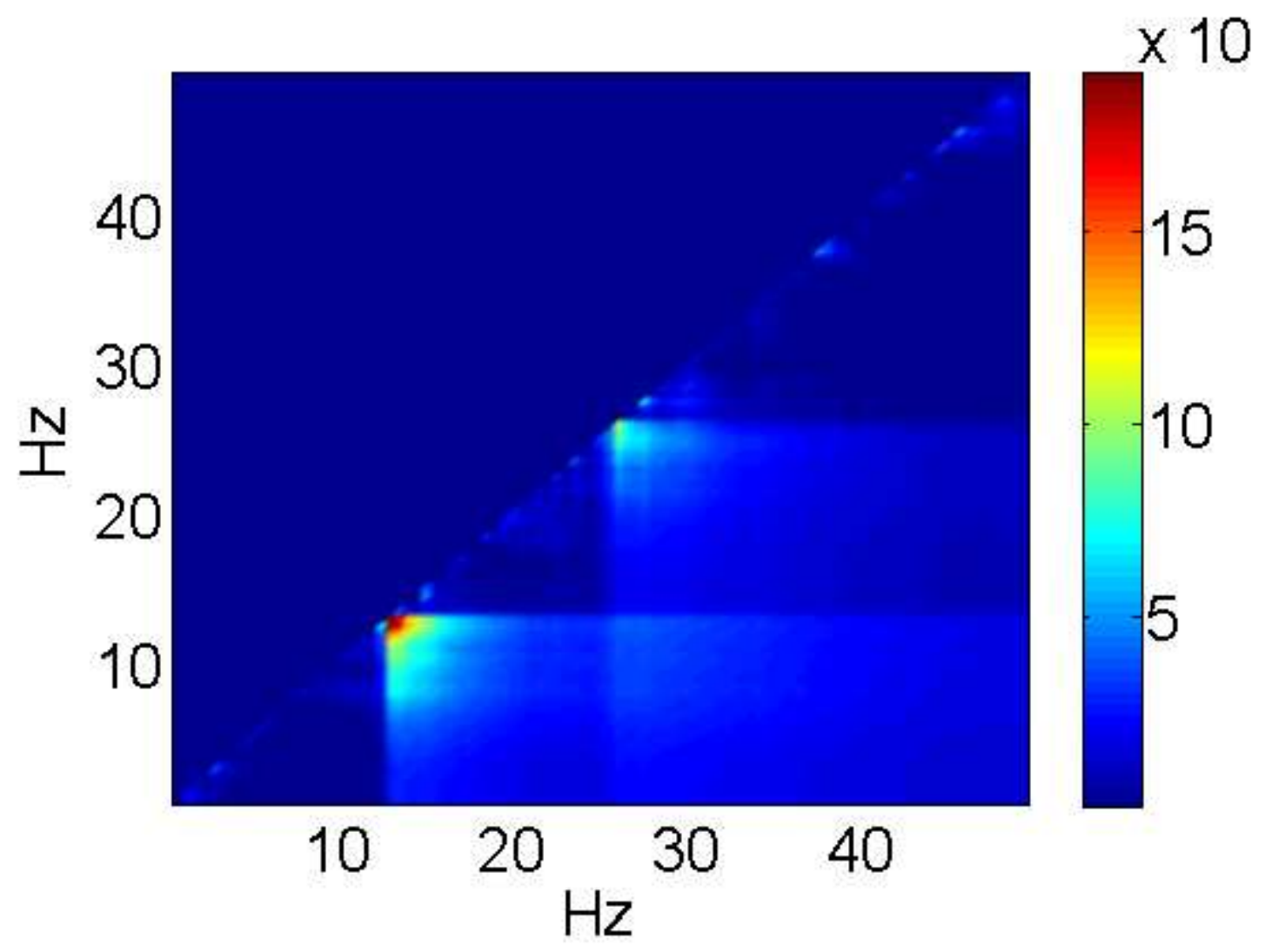}
\includegraphics[width=0.4\columnwidth]{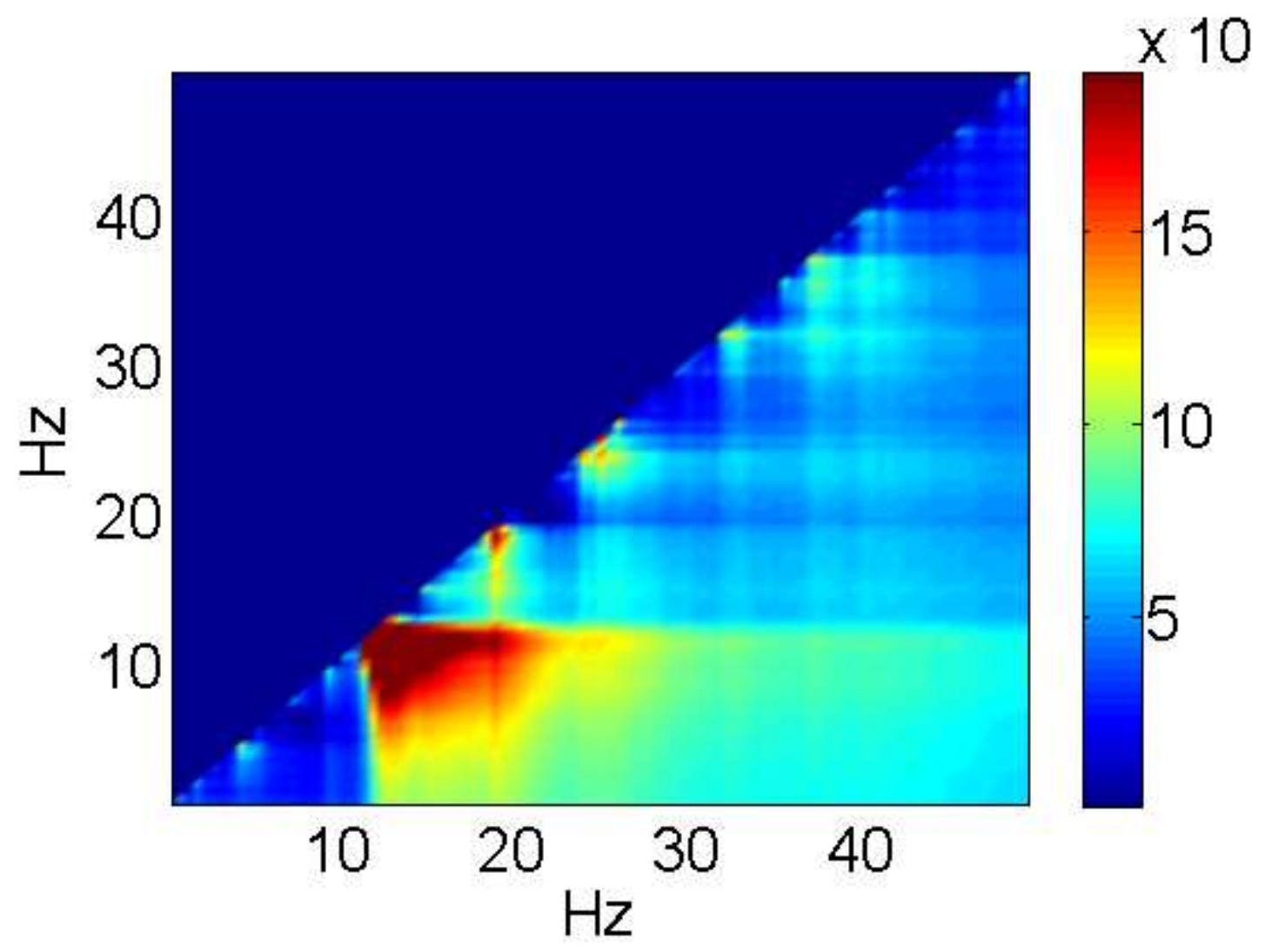}
\includegraphics[width=0.4\columnwidth]{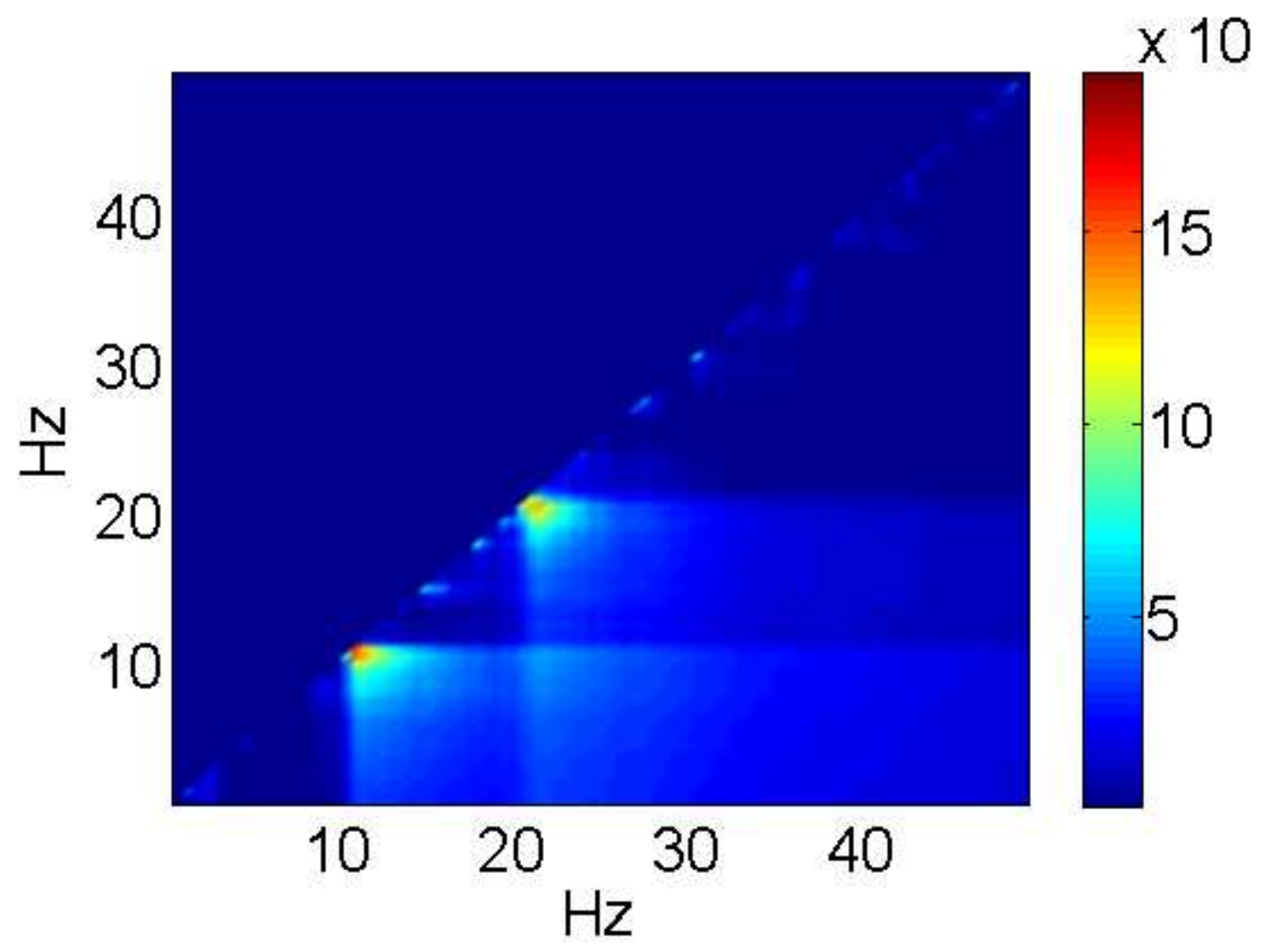}
\includegraphics[width=0.4\columnwidth]{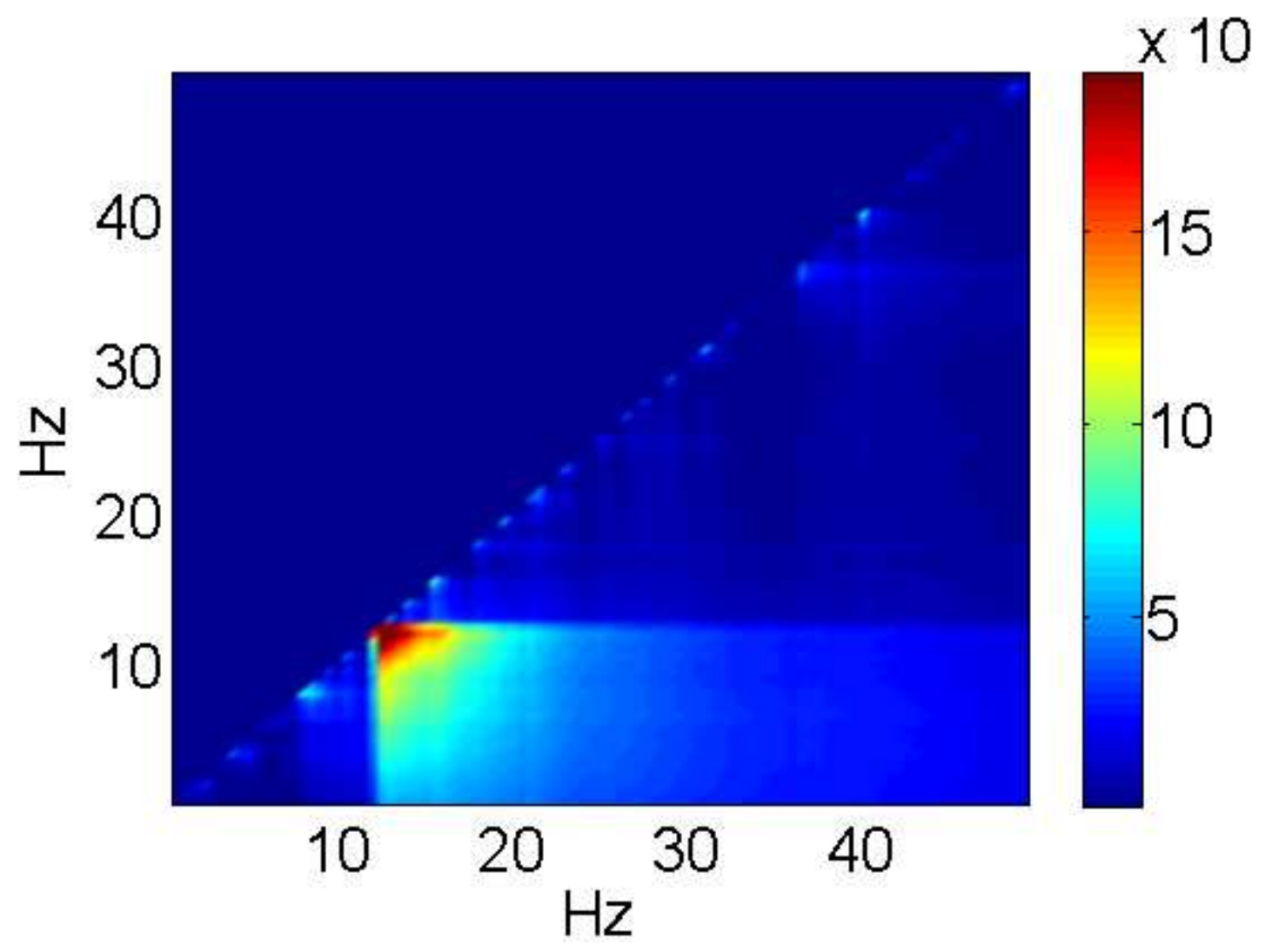}
\includegraphics[width=0.4\columnwidth]{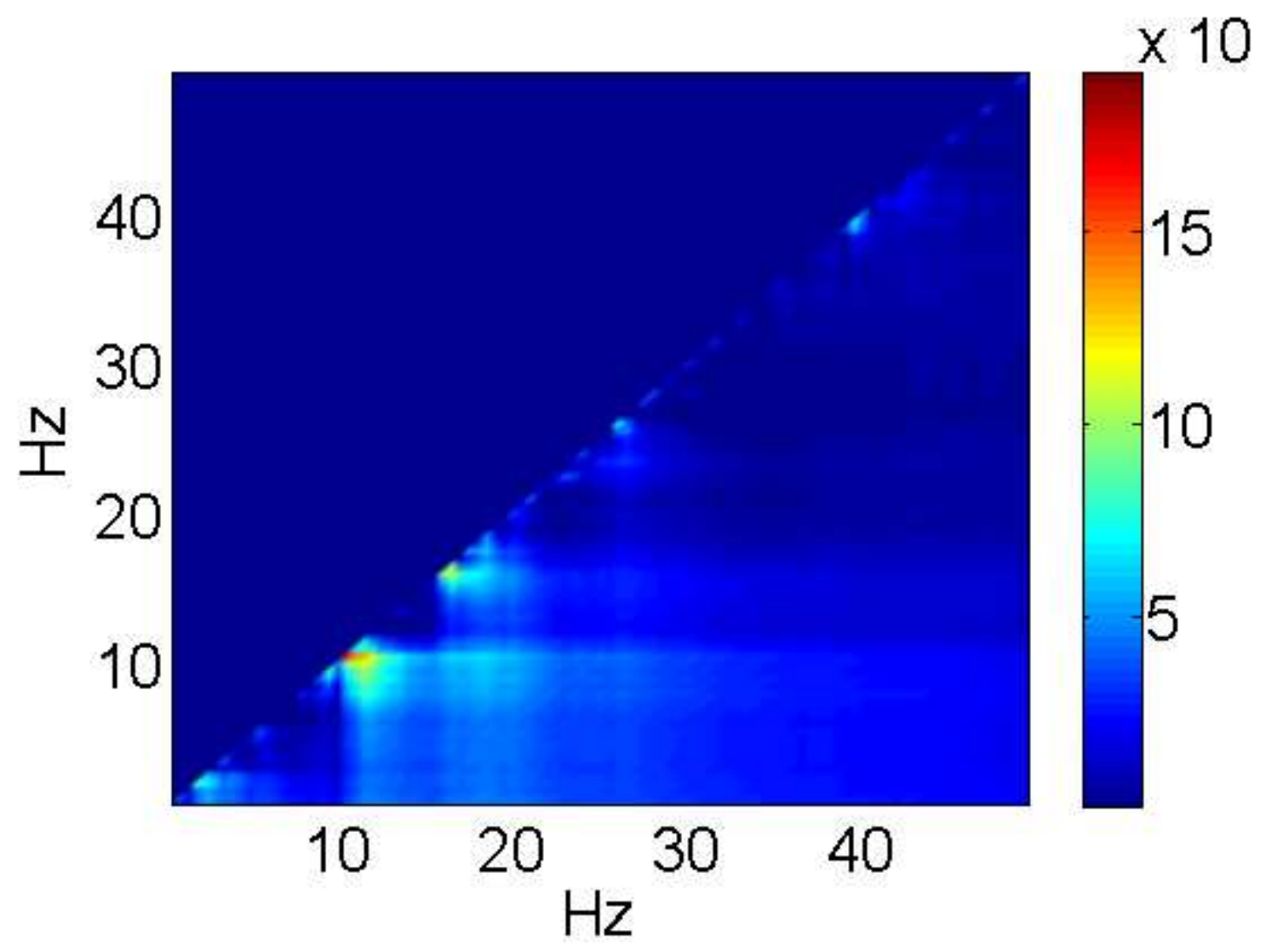}\\[-0.7cm]
\caption{HSIC, encoded by the colour value for different
  frequency bands (axes correspond to upper and lower cutoff
  frequencies). The figures, left to right, top to bottom correspond to
  subjects `aa', `al', `av', `aw' and `ay'.}
\label{fg:bci}
\end{figure*}

\vspace{-2mm}
\subsection{Brain-computer interface dataset}
\vspace{-2mm}
In this experiment, we show that BAHSIC selects features that are
meaningful in practise: we use BAHSIC to select
a frequency band for a brain-computer interface (BCI) data set
from the Berlin BCI group~\citep{DorBlaCur04}. The data
contains EEG signals (118 channels, sampled at 100 Hz) from five
healthy subjects (`aa', `al', `av', `aw' and `ay') recorded during
two types of motor imaginations. The task is to classify the
imagination for individual trials.

Our experiment proceeded in 3 steps: (\emph{i}) A Fast Fourier
transformation (FFT) was performed on each channel and the power
spectrum was computed. (\emph{ii}) The power spectra from all
channels were averaged to obtain a single spectrum for each trial.
(\emph{iii}) BAHSIC was used to select the top 5 discriminative
frequency components based on the power spectrum. The 5 selected
frequencies and their 4 nearest neighbours were used to reconstruct
the temporal signals (with all other Fourier coefficients eliminated).
The result was then passed to a normal CSP method \citep{DorBlaCur04} for
feature extraction, and then classified using a linear SVM.

We  compared automatic filtering using BAHSIC  to other
filtering approaches: normal CSP method with manual filtering
(8-40 Hz), the CSSP method \citep{LemBlaCuretal05}, and the CSSSP
method \citep{DorBlaKra05}. All results presented in
Table~\ref{tb:bci} are obtained using $50\times2$-fold
cross-validation. Our method is very competitive and obtains the
first and second place for 4 of the 5 subjects. While the CSSP and
the CSSSP methods are \emph{specialised} embedded methods (w.r.t. the CSP method) for
frequency selection on BCI data, our method is entirely generic: BAHSIC
decouples feature selection from CSP.

In Figure~\ref{fg:bci}, we use HSIC to visualise the
responsiveness of different frequency bands to motor imagination.
The horizontal and the vertical axes in each subfigure represent
the lower and upper bounds for a frequency band, respectively.
HSIC is computed for each of these bands. \cite{DorBlaKra05}
report that the $\mu$ rhythm (approx.~12 Hz) of EEG is most
responsive to motor imagination, and that the $\beta$ rhythm
(approx.~22 Hz) is also responsive. We expect that HSIC will
create a strong peak at the $\mu$ rhythm and a weaker peak at the
$\beta$ rhythm, and the absence of other responsive frequency
components will create block patterns. Both predictions are
confirmed in Figure {\ref{fg:bci}. Furthermore, the large area of
the red region for subject `al' indicates good responsiveness of
his $\mu$ rhythm. This also corresponds well with the lowest
classification error obtained for him in Table \ref{tb:bci}.


\vspace{-2mm}
\section{Conclusion}
\vspace{-2mm}

This paper proposes a backward elimination procedure for feature
selection using the Hilbert-Schmidt Independence Criterion (HSIC). The
idea behind the resulting algorithm, BAHSIC, is to choose the feature subset
that maximises the dependence between the data and labels.
With this interpretation, BAHSIC provides a unified feature selection framework
 for any form of supervised learning.
The absence of bias and good convergence properties of the
empirical HSIC estimate provide a strong theoretical jutification
for using HSIC in this context.
Although BAHSIC is a
filter method, it still demonstrates good performance compared with more specialised
methods
 in both artificial and real world data. It is also very
competitive in terms of runtime performance.\footnote{Code is freely
  available as part of the Elefant package at
  \url{http://elefant.developer.nicta.com.au}.}

\paragraph{Acknowledgments}
NICTA is funded through the Australian Government's \emph{Baking
Australia's Ability} initiative, in part through the ARC.
 This research was supported by the Pascal
Network (IST-2002-506778).

\section*{Appendix}
\begin{proof}[{\bfseries Theorem~\ref{th:unbias}}]
  Recall that $\Kb_{ii} = \Lb_{ii} = 0$. We prove the
  claim by constructing unbiased estimators for each term in
  \eq{eq:def_hsic}. Note that we have three types of expectations,
  namely $\EE_{xy} \EE_{x'y'}$, a partially decoupled expectation
  $\EE_{xy} \EE_{x'} \EE_{y'}$, and $\EE_{x} \EE_{y} \EE_{x'} \EE_{y'}$,
  which takes all four expectations independently.

  If we want to replace the expectations by empirical averages, we need
  to take care to avoid using the same discrete indices more than once
  for independent random variables. In other words, when taking
  expectations over $r$ independent random variables, we need $r$-tuples
  of indices where each index occurs exactly once. The sets
  $\ib_r^m$ satisfy this property. Their cardinalities are given
  by the Pochhammer symbols $(m)_r$. Jointly drawn random variables, on
  the other hand, share the same index. We have
  \begin{align*}
   \EE_{xy} \EE_{x'y'} \sbr{k(x,x') l(y,y')}
   = & \EE_Z \Bigl[(m)_2^{-1} \sum_{(i,j) \in \mathbf{i}_2^m} \Kb_{ij}
   \Lb_{ij}\Bigr] \\
   = & \EE_Z \sbr{(m)_2^{-1} \tr \Kb \Lb}.
  \end{align*}
  In the case of the expectation over three independent terms $\EE_{xy}
  \EE_{x'} \EE_{y'}$ we obtain
  \begin{align*}
    \EE_Z \Bigl[(m)_3^{-1}\hspace{-4mm}\sum_{(i,j,q) \in \mathbf{i}_3^m} \Kb_{ij}
   \Lb_{iq}\Bigr] = \EE_Z \sbr{(m)_3^{-1} \one^\top \Kb \Lb \one - \tr
   \Kb \Lb}.
  \end{align*}
  For four independent random variables $\EE_x \EE_y \EE_{x'} \EE_{y'}$,
  \begin{align*}
    & \EE_Z \Bigl[(m)_4^{-1}\hspace{-4mm}\sum_{(i,j,q,r) \in \mathbf{i}_4^m} \Kb_{ij}
    \Lb_{qr}\Bigr] \\
    = & \EE_Z \sbr{(m)_4^{-1} \rbr{\one^\top \Kb \one \one^\top \Lb \one
    - 4 \one^\top \Kb \Lb \one + 2 \tr \Kb \Lb}}.
  \end{align*}
  To obtain an expression for $\hsic$ we only need to take linear
  combinations using \eq{eq:def_hsic}. Collecting terms related to $\tr
  \Kb \Lb$, $\one^\top \Kb \Lb \one$, and $\one^\top \Kb \one \one^\top
  \Lb \one$ yields
  \begin{align*}
    & \hsic(\Fcal, \Gcal, \Pr_{xy}) \\
    & =
    \smallfrac{1}{m(m-3)} \EE_Z \sbr{\tr \Kb \Lb +
      \smallfrac{\one^\top \Kb \one \one^\top \Lb \one}{(m-1)(m-2)}  -
      \smallfrac{2}{m-2} \one^\top \Kb \Lb \one}.
  \end{align*}
  This is the expected value of $\hsic[\Fcal, \Gcal, Z]$.
\end{proof}
%

%
%
\vspace{-2mm}
\vspace{-3mm} \begin{proof}[{\bfseries Theorem~\ref{th:mmdkta}}]
We first relate a biased estimator of HSIC to the biased estimator
of MMD. The former is given by
\begin{align*}
  \smallfrac{1}{(m-1)^{2}} \tr \mathbf{KHLH} \text{ where }
  \Hb=\Ib-m^{-1}\one\one^\top
\end{align*}
and the bias is bounded by $O(m^{-1})$, as shown by
\cite{GreBouSmoetal05}. An estimator of MMD with bias $O(m^{-1})$
is
\begin{align*}
{\rm MMD}[\Fcal, Z] = &
\frac{1}{m_+^2}\sum_{i,j}^{m_+}k(\mathbf{x}_i,\mathbf{x}_j)+\frac{1}{m_-^2}\sum_{i,
  j}^{m_-}k(\mathbf{x}_i,\mathbf{x}_j) \\
&
-\frac{2}{m_+m_-}\sum_{i}^{m_+}\sum_{j}^{m_-}k(\mathbf{x}_i,\mathbf{x}_j)
= \tr \Kb \Lb .
\end{align*}
If we choose  $l(y,y') = \rho(y) \rho(y')$ with $\rho(1) =
m_+^{-1}$ and $\rho(-1) = m_-^{-1}$, we can see $\Lb \one =
0$. In this case $\tr \Kb \Hb \Lb \Hb = \tr \Kb \Lb$, which shows
that the biased estimators of MMD and HSIC are identical up to a
constant factor. Since the bias of $\tr \Kb \Hb \Lb \Hb$ is
$O(m^{-1})$, this implies the same bias for the MMD estimate.

To see the same result for Kernel Target Alignment, note that for equal
class size the normalisations with regard to $m_+$ and $m_-$ become
irrelevant, which yields the corresponding MMD term.
\end{proof}



\vspace{-2mm}
\small
\bibliographystyle{natmlapa}
\bibliography{shortBib}


\end{document}